\documentclass[11pt,a4paper]{article}
\usepackage[hyperref]{emnlp-ijcnlp-2019}
\usepackage{times}
\usepackage{latexsym}

\usepackage{url}
\usepackage{helvet}
\usepackage{courier}
\usepackage{stfloats}
\usepackage{color}
\usepackage{amsmath}
\usepackage{amssymb}
\usepackage{array}
\usepackage{graphicx}
\usepackage{xcolor}
\usepackage{pgfplots}
\usepackage{subfigure}
\usepackage{colortbl} 
\usepackage{arydshln} 
\usepackage{multirow} 
\usepackage{multicol}
\usetikzlibrary{arrows,automata}
\usepackage{tikz}
\usepackage{float}
\usepackage{mathrsfs}
\usepackage{mathtools}
\usepackage{amsopn}
\usepackage{bm}
\usepackage{soul}

\aclfinalcopy 

\title{Syntax-Directed Attention for Neural Machine Translation}
\author{Kehai Chen$^1$\thanks{\ Kehai Chen was an internship research fellow at NICT when conducting this work.}, Rui Wang$^2$\thanks{\ Corresponding author.}, Masao Utiyama$^2$, Eiichiro Sumita$^2$, Tiejun Zhao$^1$ \\
	$^1$Harbin Institute of Technology, Harbin, China \\
	$^2$National Institute of Information and Communications Technology, Kyoto, Japan \\
	\{khchen, tjzhao\}@hit.edu.cn, \{wangrui, mutiyama, eiichiro.sumita\}@nict.go.jp}
\date{}

\begin{document}
\maketitle
\begin{abstract}
\emph{Attention mechanism}, including global attention and local attention, plays a key role in neural machine translation (\textbf{NMT}).
Global attention attends to all source words for word prediction.
In comparison, local attention selectively looks at fixed-window source words.
However, alignment weights for the current target word often decrease to the left and right by linear distance centering on the aligned source position and neglect syntax distance constraints.
In this paper, we extend the local attention with syntax-distance constraint, which focuses on syntactically related source words with the predicted target word to learning a more effective context vector for predicting translation.
Moreover, we further propose a double context NMT architecture, which consists of a global context vector and a syntax-directed context vector from the global attention, to provide more translation performance for NMT from source representation.
The experiments on the large-scale Chinese-to-English and English-to-German translation tasks show that the proposed approach achieves a substantial and significant improvement over the baseline system.
\end{abstract}

\section{Introduction}
\label{Intro}
Recent works of neural machine translation (\textbf{NMT}) have been proposed to adopt the encoder-decoder framework \cite{kalchbrenner-blunsom:2013:EMNLP,cho-EtAl:2014:SSST-8,Sutskever:2014:SSL:2969033.2969173}, which employs a recurrent neural network (\textbf{RNN}) encoder to represent source sentence and a RNN decoder to generate target translation word by word.
Especially, the NMT with an \emph{attention mechanism} (called as global attention) is proposed to acquire source sentence context dynamically at each decoding step, thus improving the performance of NMT~\cite{Bahdanau-EtAl:2015:ICLR2015}.
The global attention is further refined into a local attention~\cite{luong-pham-manning:2015:EMNLP}, which selectively looks at fixed-window source context at each decoding step, thus demonstrating its effectiveness on WMT translation tasks between English and German in both directions.

\begin{figure*}[thb!]
	\centering
	\subfigure{
		\begin{minipage}[b]{0.5\linewidth}
			\includegraphics[width=1.0\textwidth]{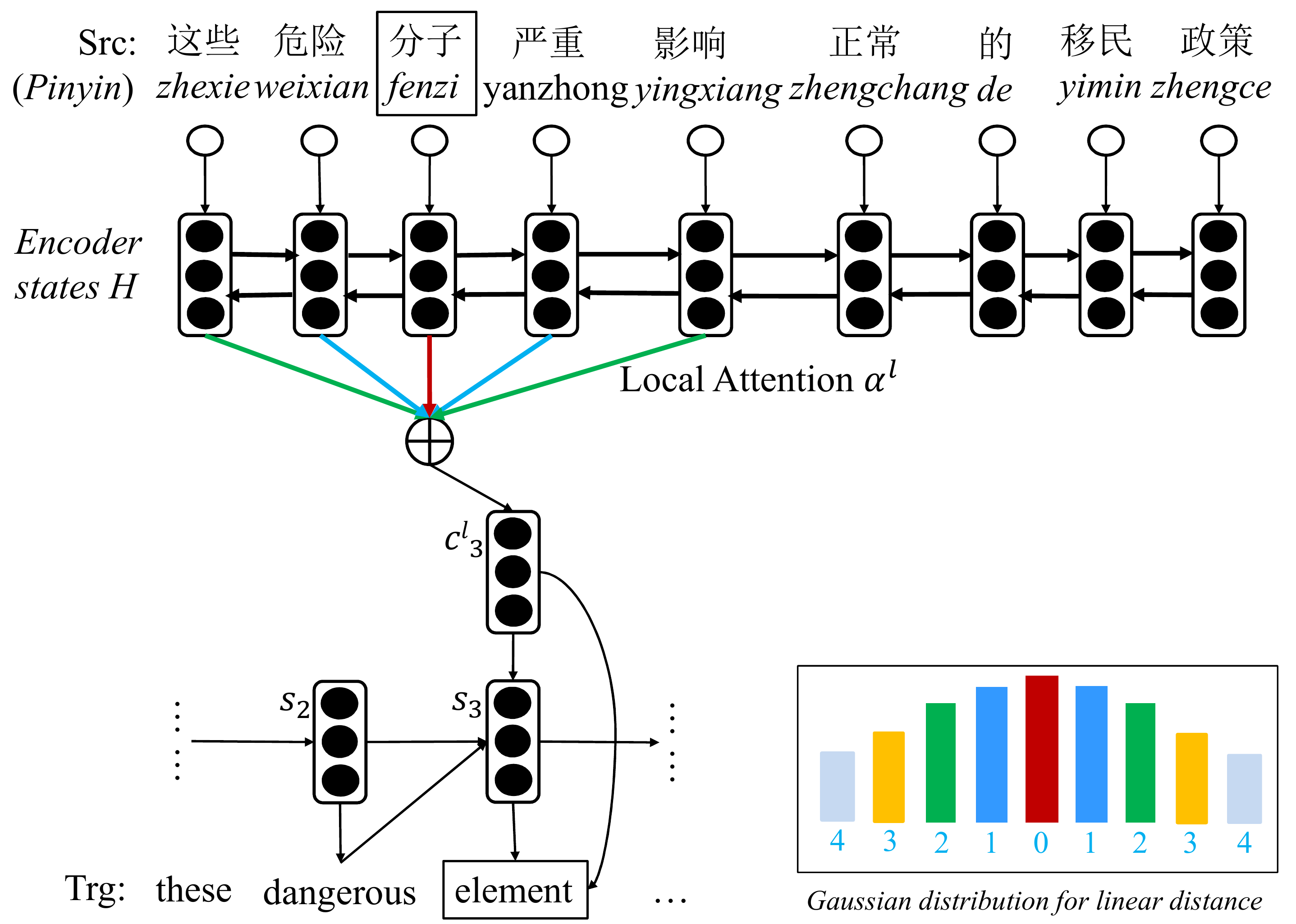}
			\centerline{(a)}
		\end{minipage}
	}
	\subfigure{
		\begin{minipage}[b]{0.38\linewidth}
			\centering
			\includegraphics[width=1.0\textwidth]{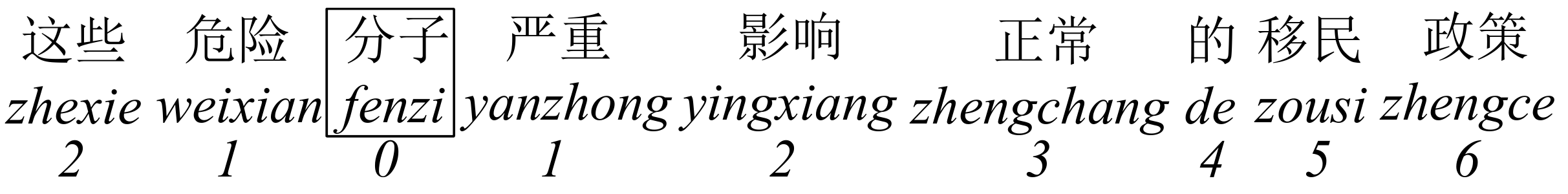}
			\centerline{(b)}
			\centering
			\includegraphics[width=0.7\textwidth]{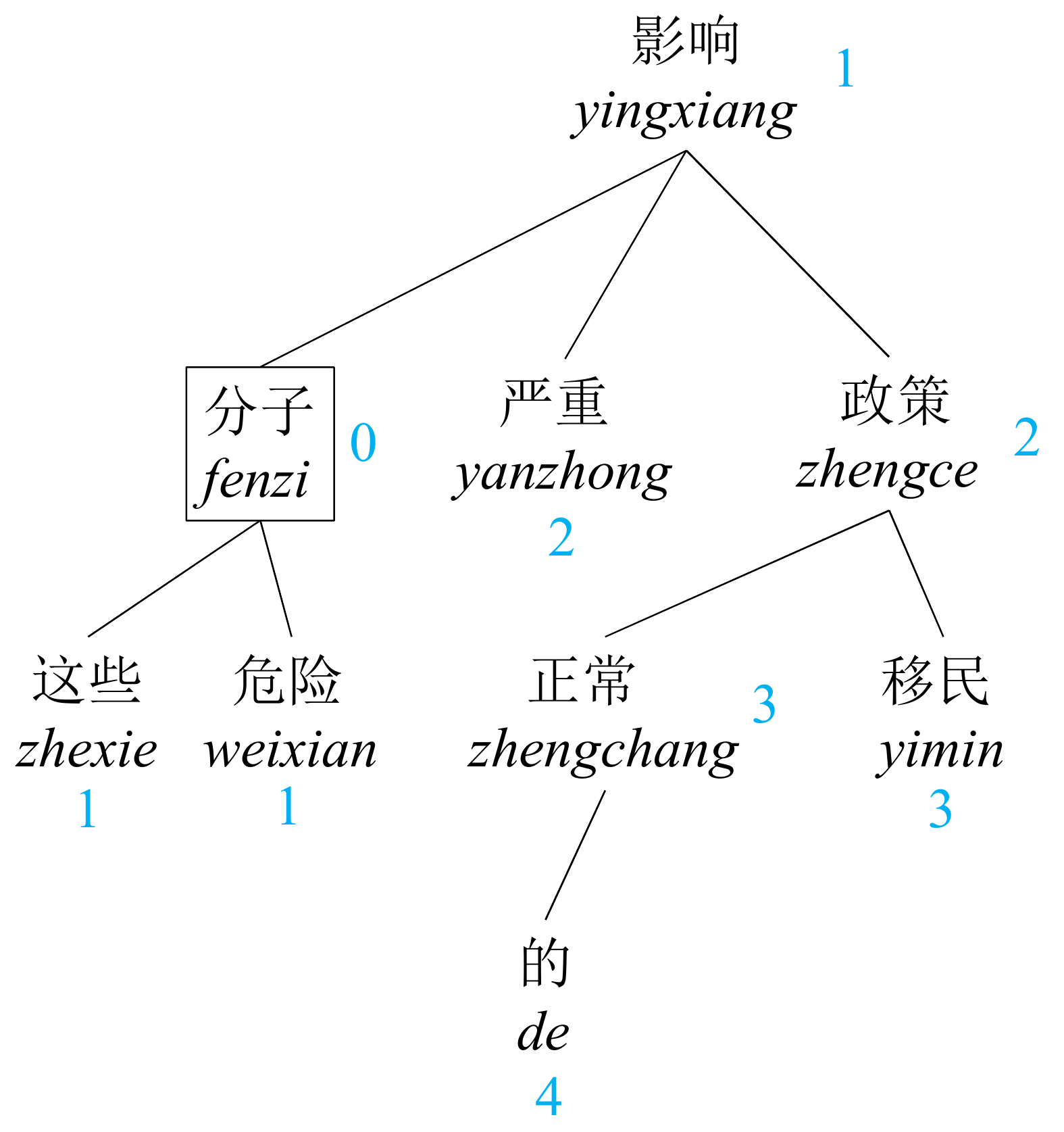}
			\centerline{(c)}
		\end{minipage}
	}
	\caption{(a) NMT with the local attention. The black dotted box is the current source aligned word and the red dotted box is the predicted target word. (b) Linear distances for the source word ``\emph{fenzi}", for which the number denotes the linear distance. (c) Syntax-directed distances for source word ``\emph{fenzi}", for which the blue number represents syntax-directed distance between each word and ``\emph{fenzi}".}
	\label{fig:motavation}
\end{figure*}

Specifically, the local attention first predicts a single aligned source position $p_{i}$ for the current time-step $i$.
The decoder focuses on the fixed-window encoder states centered around the source position $p_{i}$, and compute a context vector $\textbf{\emph{c}}^{l}_{i}$ by alignment weights $\alpha^{l}$ for predicting current target word.
Figure~\ref{fig:motavation}(a) shows a Chinese-to-English NMT model with the local attention, and its contextual window is set to five.
When the aligned source word is ``\emph{fenzi}", the local attention focuses on source words \{``\emph{zhexie}", ``\emph{weixian}", ``\emph{fenzi}", ``\emph{yanzhong}", ``\emph{yingxiang}"\} in the window to compute its context vector.
Meanwhile, the local attention is to obtain the positions of five encoder states by Gaussian distribution, which penalty their alignment weights according to the distance with word ``\emph{fenzi}".
For example, the syntax distances of these five source words are \{\emph{2}, \emph{1}, \emph{0}, \emph{1}, \emph{2}\} in contextual window, as shown in Figure~\ref{fig:motavation}(b).
In other words, the greater the distance from the aligned word in the window is, the smaller the source words in the window to the context vector would contribute.
In spite of its success, the local attention is to encode source context and compute a local context vector by linear distance centered around current aligned source position.
It does not take syntax distance constraints into account.

Figure~\ref{fig:motavation}(c) shows the dependency tree of the Chinese sentence in Figure~\ref{fig:motavation}(b).
Support the word  ``\emph{fenzi}$_{0}$" as the aligned source word, its syntax-distance neighbor window is \{``\emph{zhexie}$_1$", ``\emph{weixian}$_{1}$", ``\emph{fenzi}$_{0}$", ``\emph{yingxiang}$_{1}$", ``\emph{yanzhong}$_{2}$", ``\emph{zhengce}$_{2}$''\} , where the footnote of a word is its syntax-distance with the central word.  In comparison,  its local neighbor window is  \{``\emph{zhexie}", ``\emph{weixian}", ``\emph{yanzhong}", ``\emph{yingxiang}", ``\emph{zhengchang}"\} based on linear distance. Note that the ``\emph{zhengce}" is very informative for the correct translation, but it is far away from ``\emph{fenzi}" such that it is not easy to be focused by the local attention.
Besides, the syntax distances of ``\emph{yanzhong}" and ``\emph{yingxiang}" are $two$ and $one$, but the linear distances are $one$ and $two$.
This means that the ``\emph{yingxiang}'' is syntactically more relevant to the ``\emph{fenzi}" than ``\emph{yingxiang}".
However, the existing \emph{attention mechanism}, including the global or local attention, does not allow NMT to distinguish  syntax distance constraint from source representation.

In this paper, we extend the local attention with a novel syntax-distance constraint, to capture syntax related source words with the predicted target word.
Following the dependency tree of a source sentence, each source word has a syntax-distance constraint mask, which denotes its syntax distance with the other source words.
The decoder then focuses on the syntax-related source words within the syntax-distance constraint to compute a more effective context vector for predicting target word.
Moreover, we further propose a \emph{double context} NMT architecture, which consists of a global context vector and a syntax-directed local context vector from the global attention, to provide more translation performance for NMT from source representation.
The experiments on the large-scale Chinese-to-English and English-to-German translation tasks show that the proposed approach achieves a substantial and significant improvement over the baseline system.

\section{Background}
\label{Background}
\subsection{Global Attention-based NMT}
\label{GAttNMT}
In NMT~\cite{Bahdanau-EtAl:2015:ICLR2015}, the context of translation prediction relies heavily on \emph{attention mechanism} and source input. 
Typically, the decoder computes a alignment score $e_{ij}$ between each source annotation $\textbf{\emph{h}}_j$ and predicted target word $y_i$ according to the previous decoder hidden state $\textbf{\emph{s}}_{i-1}$
\begin{equation}
e_{ij} = f(\textbf{\emph{s}}_{i-1}, \textbf{\emph{h}}_{j}),
\label{eq:AlignmentWeight}
\end{equation}
where $f$ is a RNN with GRU.
Then all alignment scores are normalized to compute weight $\alpha_{ij}$ of each encoder state $\textbf{\emph{h}}_j$
\begin{equation}
\alpha_{ij} = \frac{exp(e_{ij})}{\sum^{J}_{k=1}exp(e_{ik})}.
\label{eq:GlobalAttention}
\end{equation}
Furthermore, the $\alpha_{ij}$ is used to weight all source annotations for computing current time-step context vector $\textbf{\emph{c}}^{g}_i$:
\begin{equation}
\textbf{\emph{c}}^{g}_i = \sum_{j=1}^{J}\alpha_{ij}\textbf{\emph{h}}_{j}.
\label{eq:GlobalContextVector}
\end{equation}

Finally, the context vector $\textbf{\emph{c}}_i$ is used to predict target word $y_i$ by a non-linear layer:
\begin{multline}
P(y_i|y_{ < i},x) = \\
softmax(\textbf{\emph{L}}_{o}\textbf{tanh}(\textbf{\emph{L}}_{w}\textbf{\emph{E}}_{y}[\hat{y}_{i-1}]+\textbf{\emph{L}}_{d}\textbf{s}_i+\textbf{\emph{L}}_{cg}\textbf{c}^{g}_{i}))
\label{eq:GlobalPrediction}
\end{multline}
where $\textbf{\emph{s}}_{i}$ is the current decoder hidden state and $y_{i-1}$ is the previously emitted word; the matrices $\textbf{\emph{L}}_{o}$, $\textbf{\emph{L}}_{w}$,
$\textbf{\emph{L}}_{d}$ and $\textbf{\emph{L}}_{cg}$ are transformation matrices.
Intuitively, this attention is called as \emph{global attention} because of the context vector $\textbf{c}^{g}_i$ takes all source words into consideration~\cite{luong-pham-manning:2015:EMNLP}.

\subsection{Local Attention-based NMT}
\label{LAttNMT}
Compared with the \emph{global attention}, the \emph{local attention} selectively focuses on a small window of context~\cite{luong-pham-manning:2015:EMNLP}.
It first generates a source aligned position $p_i$ for the predicted target word at current decoder time-step \emph{i}:
\begin{equation}
p_{i} =J \cdot sigmoid(\textbf{\emph{v}}^{T}tanh(\textbf{\emph{W}}_{p}\textbf{\emph{h}}^{'}_{i})),
\label{eq:AlignedPosition}
\end{equation}
where $J$ is the length of source sentence and $\textbf{\emph{h}}^{'}_i$ is decoder hidden state, $\textbf{\emph{v}}^{T}$ and $\textbf{\emph{W}}_{p}$ are weights.

To focus on source words within the fixed-window, the $\alpha^{l}_{ij}$ is refined by the follow eq.\eqref{eq:LocalAttention}:
\begin{equation}
\alpha^{l}_{ij} =
\begin{cases}
\alpha_{ij}exp(-\frac{(j-p_i)^{2}}{2\sigma^{2}}), &j \in [p_i-D, p_i+D] \cr 0, &j \notin [p_i-D, p_i+D],
\end{cases}
\label{eq:LocalAttention}
\end{equation}
where [$p_i$-\emph{D}, $p_i$+\emph{D}] denotes the local window and the standard deviation is empirically set as $\sigma = \frac{D}{2}$.\footnote{The \emph{D} is set as \emph{10} in local attention of \cite{luong-pham-manning:2015:EMNLP}.}
Moreover, the local attention focuses on source annotations in window [$p_i-D, p_i+D$] to compute the current time-step local context vector $\textbf{\emph{c}}^{l}_i $:
\begin{equation}
\textbf{\emph{c}}^{l}_i = \sum_{j\in [p_i-D, p_i+D]}\alpha^{l}_{ij}\textbf{\emph{h}}_{j}.
\label{eq:LocalContextVector}
\end{equation}

Finally, the context vector $\textbf{\emph{c}}^{l}_i$ is then used to predict target word $y_i$ by a non-linear layer:
\begin{multline}
P(y_i|y_{ < i}, x) = \\
softmax(\textbf{\emph{L}}_{o}\textbf{tanh}(\textbf{\emph{L}}_{w}\textbf{\emph{E}}_{y}[\hat{y}_{i-1}]+\textbf{\emph{L}}_{d}\textbf{s}_i+\textbf{\emph{L}}_{cl}\textbf{\emph{c}}^{l}_{i})),
\label{eq:LocalPrediction}
\end{multline}
where $\textbf{\emph{s}}_{i}$ is the current decoder hidden state and $y_{i-1}$ is the previously emitted target word.

\section{Syntax-Directed Attention}
\label{SAttNMT}
\subsection{Syntax Distance Constraint}
\label{SDConstraint}
In NMT, the decoder computes the current context vector by weighting each encoder state with alignment weight to predict target word.
Actually, these alignment weights are defined by the linear distance with the aligned source center position, such as the word ``\emph{fenzi}" in Figure~\ref{fig:motavation}(a).
In other words, the greater the distance to the center position is, the smaller the contribution of the source word to the context vector is.
Recently, the source long-distance dependency has been explicitly explored to enhance the encoder of NMT, thus improving target word prediction~\cite{Chen-EtAl:2017:EMNLP2017,ijcai2017-584}.
This means that syntax context is beneficial for NMT.
However, the existing NMT cannot adequately capture the source syntax context by the linear distance attention mechanism.

\begin{figure}[thb!]
	\centering
	\includegraphics[width=0.46\textwidth]{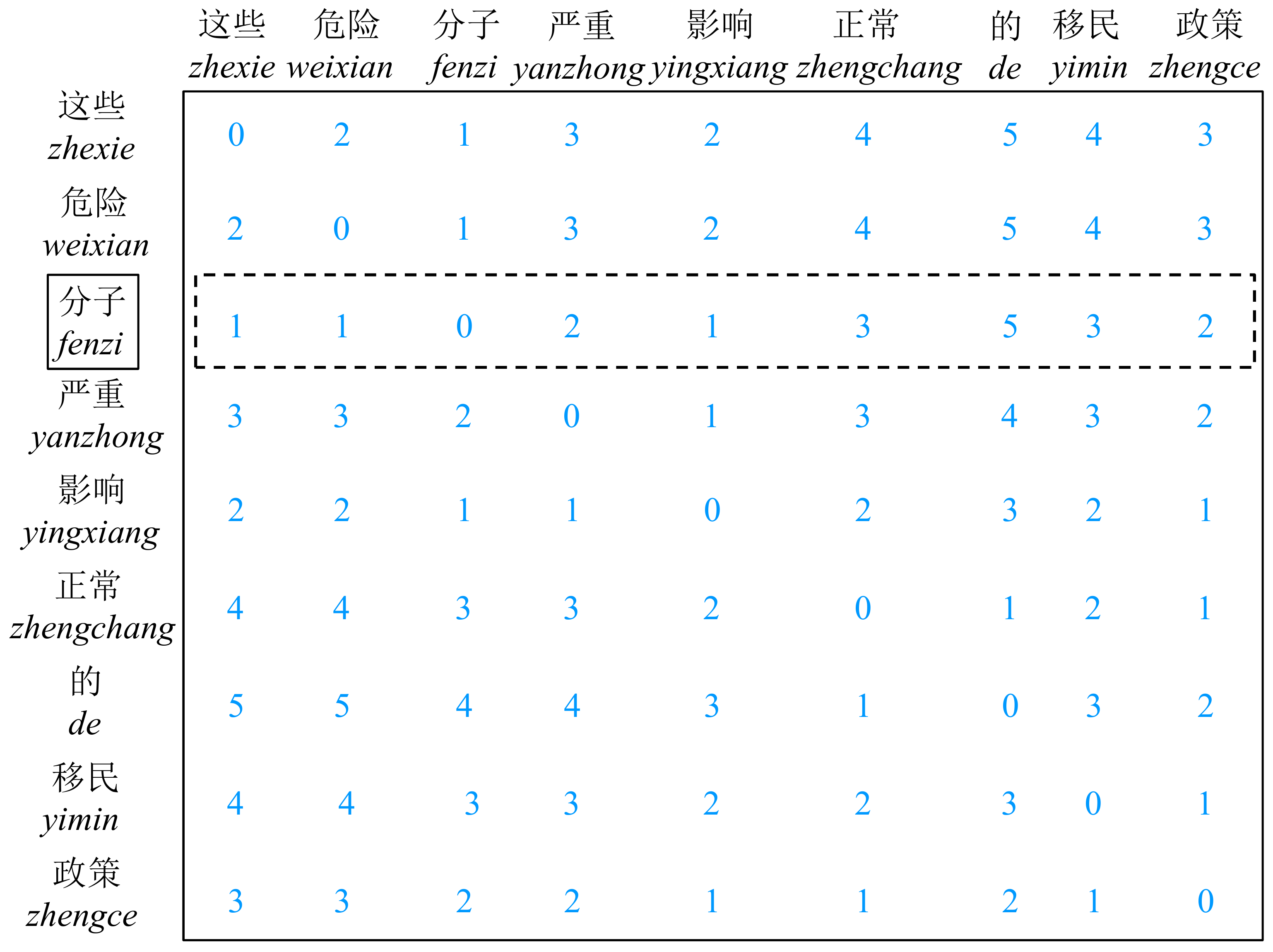}
	\caption{Syntax distance constraint mask matrix $\mathcal{M}$ for the dependency-based Chinese sentence in Figure~\ref{fig:motavation}(c), in which each row denotes the syntax distance mask of one source word, for example the dotted black box is syntax distance constraint mask for source word ``\emph{fenzi}".}
	\label{fig:SCMMatrix}
\end{figure}

To address this issue, we propose a syntax distance constraint (\textbf{SDC}), in which we learn a SDC mask for each source word, as shown in Figure~\ref{fig:SCMMatrix}.
Specifically, given a source sentence $F$ with dependency tree $T$, each node denotes a source word $x_j$ and the distance between two connected nodes is defined as $one$.
We then traverse every word according to the order of source word, and compute the distances of all remaining words to the current traversed word $x_j$ as its SDC mask $m_j$.
Finally, we learn a sequence of SDC mask \{$m_0, m_1, ..., m_J$\}, and organize them as a $J*J$ matrix $\mathcal{M}$, in which $J$ denotes the length of source sentence, and elements in each row denote the distances of all word to the row-index word,
\begin{equation}
\mathcal{M} = [[m_0],[m_1], ..., [m_J]].
\label{eq:SCMMatrix}
\end{equation}

As shown in Figure~\ref{fig:SCMMatrix}, the third row denotes the syntax context mask of word ``\emph{fenzi}".
Specifically, syntax distance of ``\emph{fenzi}" itself is zero; the syntax distances of ``\emph{zhexie}", ``\emph{weixian}", and ``\emph{yingxiang}" are one; the syntax distance of  ``\emph{yanzhong}" and ``\emph{zhengce}" are two; the syntax distance of ``\emph{zhengchang}" and  ``\emph{yimin}", and ``\emph{de}" are four, as shown the black dotted box in Figure~\ref{fig:SCMMatrix}.
\begin{figure*}[thb!]
	\centering
	\begin{minipage}[b]{0.48\linewidth}
		\includegraphics[width=3.1in,height=2.6in]{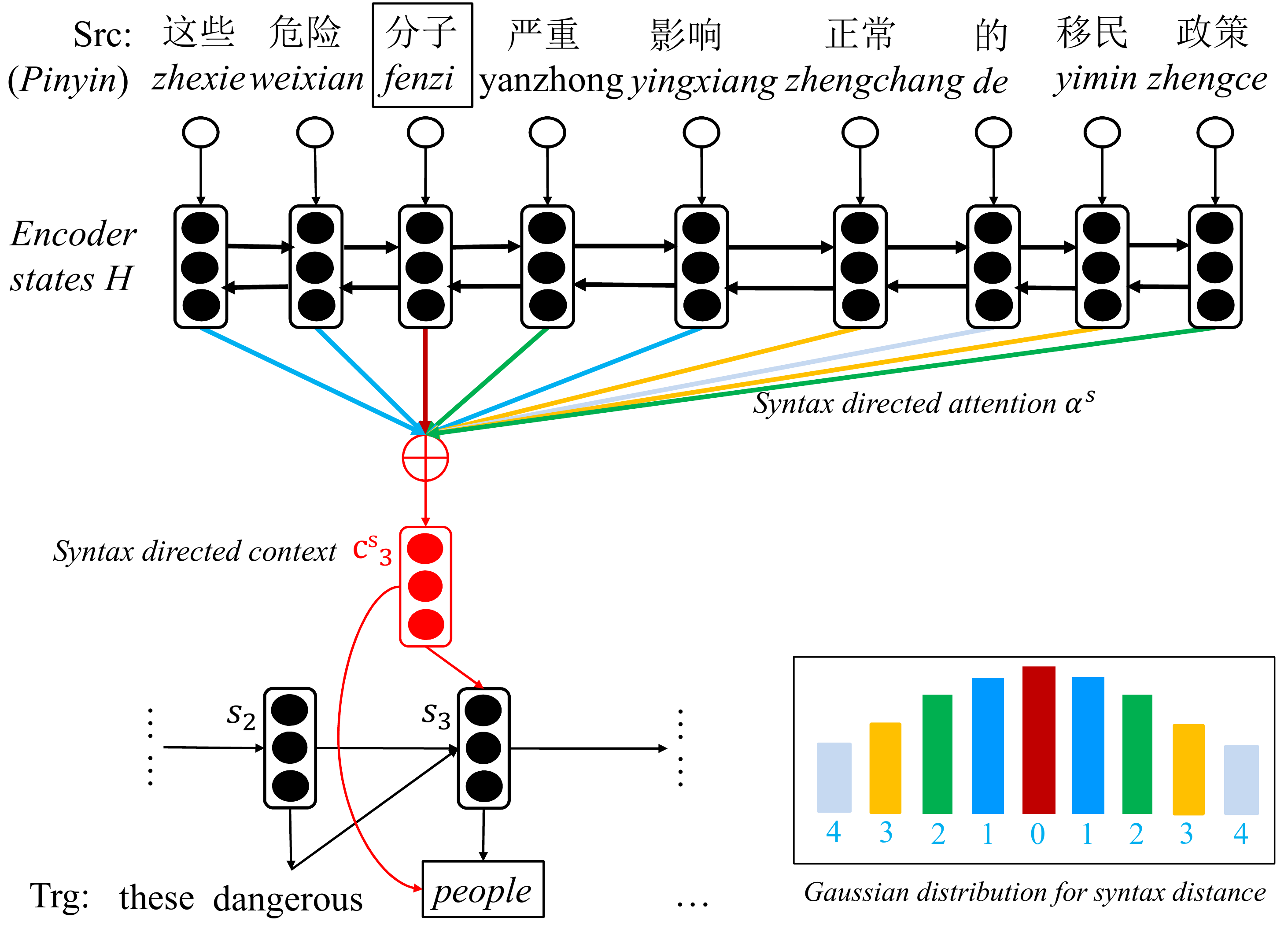}
		\caption{Syntax-directed attention for NMT.}
		\label{fig:SAttNMT}
	\end{minipage}
	\begin{minipage}[b]{0.48\linewidth}
		\includegraphics[width=3.1in,height=2.7in]{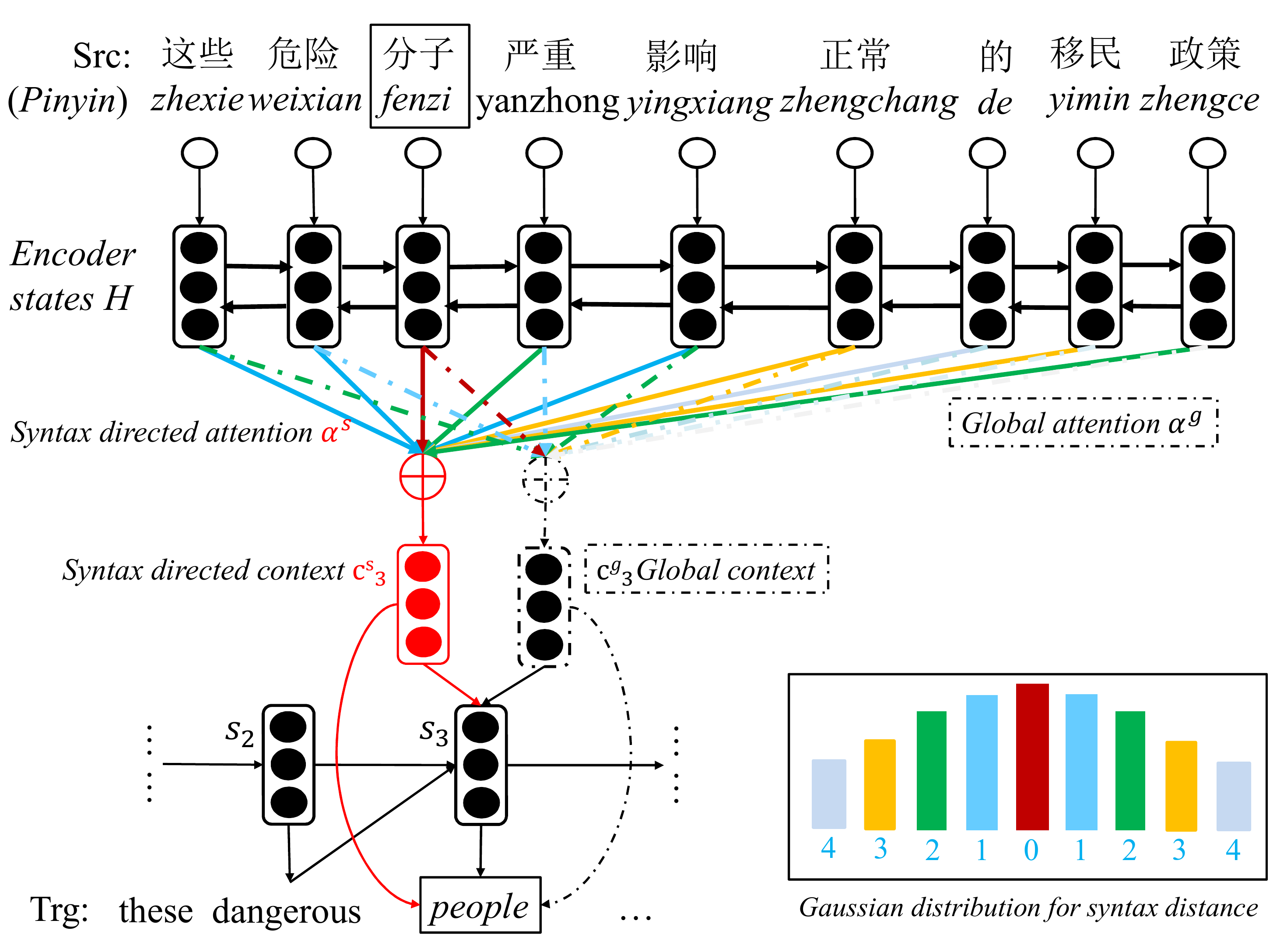}
		\caption{Double context for NMT.}
		\label{fig:DCAttNMT}
	\end{minipage}
\end{figure*}

\subsection{Syntax-Directed Attention}
\label{SANMT}
To capture the source context with the SDC (in Section~\ref{SDConstraint}), we propose a novel syntax-directed attention (\textbf{SDAtt}) for NMT, as shown in Figure~\ref{fig:SAttNMT}.
The decoder first learn aligned source position $p_i$ of the current time-step $i$ by the eq.\eqref{eq:AlignedPosition}.
According to the position $p_i$, we obtain its SDC mask $m_i$ from matrix $\mathcal{M}$ in eq.\eqref{eq:SCMMatrix}, i.e., $\mathcal{M}[p_i]$.
We learn alignment score $e^{s}_ij$ with SDC mask $\mathcal{M}[p_i]$ by the following equation:
\begin{equation}
e^{s}_{ij} = e_{ij}exp(-\frac{(\mathcal{M}[p_i][j])^{2}}{2\sigma^{2}}),
\label{eq:SytanxAlignmentWeight}
\end{equation}
where the Gaussian distribution centered around $p_i$ is used to capture the difference of syntax distance, and further to tune the alignment score $e_{ij}$ in eq.\eqref{eq:AlignmentWeight}.
Besides, the standard deviation $\sigma$ is set as $\frac{n}{2}$, in which one syntax distance is different from one linear distance of the local attention.
In other words, one syntax distance corresponds to multiple syntax-related words instead of two words in local attention.
The $n$ is more similar to the order of \emph{n}-gram language model.
Therefore, the $n$ is empirically set as four in our experiments, which means that we only take \emph{4}-gram SDC into account.

The $\alpha^{s_n}_{ij}$ is normalized within \emph{n}-gram SDC:
\begin{equation}
\alpha^{s_{n}}_{ij}  =
\begin{cases}
\frac{exp(e^{s}_{ij})}{\sum_{k\in M[p_i][k] \le n}exp(e^{s}_{ik})}, &j \in [p_{i}-n, p_{i}+n] \cr 0, & j \notin [p_i-n, p_i+n],
\end{cases}
\label{eq:LocalSyntaxAttention}
\end{equation}
In other words, we only consider words within the \emph{n}-gram SDC and simply ignore the outside part of the \emph{n}-gram SDC.

The context vector $\textbf{\emph{c}}^{s}_i$ is, then, computed as a weighted sum of these annotations $\textbf{\emph{h}}_i$ by alignment weights with the SDC:
\begin{equation}
\textbf{\emph{c}}^s_i = \sum_{j}^{J}\alpha^{s_n}_{ij}\textbf{\emph{h}}_{j},
\label{eq:SyntaxContextVector}
\end{equation}
Finally, similar to the eq.\eqref{eq:LocalPrediction}, the context vector $\textbf{\emph{c}}^s_i$ is used to predict the target word $y_i$:
\begin{multline}
P(y_i|y_{ < i}, x, T) = \\
softmax(\textbf{\emph{L}}_{o}\textbf{tanh}(\textbf{\emph{L}}_{w}\textbf{\emph{E}}_{y}[\hat{y}_{i-1}]+\textbf{\emph{L}}_{d}\textbf{\emph{s}}_i+\textbf{\emph{L}}_{cs}\textbf{\emph{c}}^{s}_{i}))
\label{eq:SAWordPrediction}
\end{multline}
where $\textbf{\emph{s}}_{i}$ is the current decoder hidden state and $y_{i-1}$ is the previously emitted word.

\subsection{Double Context Mechanism}
\label{DCNMT}
In Section~\ref{SANMT}, the proposed SDAtt uses a local context with the SDC to compute current context vector instead of context vector with linear distance constraint in global or local attention.
Inspired by the decoder with additional visual attention~\cite{calixto-liu-campbell:2017:Long,Chen-EtAl:2017:EMNLP2017}, we design a unique \textbf{\emph{double context}} NMT as shown in Figure~\ref{fig:DCAttNMT}, to provide more translation performance for NMT from SDAtt in Section~\ref{SANMT}. 
The proposed model can be seen as an expansion of the global attention NMT framework described in Section~\ref{GAttNMT} with the addition of a SDAtt to incorporate source syntax distance constraint.

Compared with the global attention, we learn two context vectors over a single global attention for target word prediction: a traditional (global) context vector which always attends to all source words and a syntax-directed context vector that focuses on \emph{n}-gram (i.e., \emph{4}-gram) source syntax context words. 
To that end, in addition to the traditional context vector $\textbf{\emph{c}}^{g}_i$ in eq.\eqref{eq:GlobalContextVector}, we learn a context vector $\textbf{\emph{c}}^{s}_i$ for the SDC according to the eq.\eqref{eq:SyntaxContextVector}.
Formally, the probability for the next target word is computed by the following eq.\eqref{eq:DCWordPrediction},

\begin{multline}
P(y_i|y_{ < i}, x, T) = softmax(\textbf{\emph{L}}_{o}\textbf{tanh}(\textbf{\emph{L}}_{w}\textbf{\emph{E}}_{y}[\hat{y}_{i-1}]+\\
\textbf{\emph{L}}_{d}\textbf{\emph{s}}_i+\textbf{\emph{L}}_{cg}\textbf{\emph{c}}^{g}_{i}+\textbf{\emph{L}}_{cs}\textbf{\emph{c}}^{s}_{i})).
\label{eq:DCWordPrediction}
\end{multline}

\section{Experiments}
\label{Experiments}
\subsection{Data sets}
The proposed methods were evaluated on two data sets.\footnote{Our method also was verified on the English-to-French translation task of the WMT'14 data set}
\begin{itemize}
	\item For English (EN) to German (DE) translation task, \emph{4.43} million bilingual sentence pairs of the WMT'14 data set was used as the training data, including Common Crawl, News Commentary and Europarl v7.
	The newstest2012 and newstest2013/2014/2015 was used as dev set and test sets, respectively.
	\item For Chinese (ZH) to English (EN) translation task, the training data set was \emph{1.42} million bilingual sentence pairs from LDC corpora, which consisted of LDC2002E18, LDC2003E07, LDC2003E14, Hansards portion of LDC2004T07, LDC2004T08, and LDC2005T06.
	The NIST02 and the NIST03/04/05/06/08 data sets were used as dev set and test sets, respectively.
\end{itemize}

\begin{table*}[htbp!]
	\centering
	\scalebox{.95}{
		\begin{tabular}{l|c|c|c|c|c|c|c}
			\hline
			\hline
			{\textbf{ZH-EN}} & {\textbf{Dev (NIST02)}} & {\textbf{NIST03}} & {\textbf{NIST04}} & {\textbf{NIST05}} & {\textbf{NIST06}} & {\textbf{NIST08}} & {\textbf{AVG}}\\
			\hline
			PBSMT                             & 33.15  & 31.02  & 33.78  & 30.33  & 29.62  & 23.53  & 29.66  \\
			\hline
			GlobalAtt                & 37.12  & 35.24  & 37.49  & 34.60  & 32.48  & 26.32  & 33.23  \\
			Chen et al. (\citeyear{Chen-EtAl:2017:EMNLP2017})    & 37.42  & 35.98  & 38.34  & 35.28  & 33.58  & 27.23  & 34.08  \\
			LocalAtt                  & 37.31  & 35.57  & 37.85  & 34.93  & 32.74  & 26.83  & 33.58  \\
			FlexAtt              & 37.19  & 35.46  & 37.81  & 34.76  & 32.83  & 26.71  & 33.51  \\
			\hline
			\textbf{SDAtt}               & \textbf{38.01}  & \textbf{36.67}$^{**\dagger}$  & \textbf{38.66}$^{**\dagger}$  & \textbf{35.74}$^{**\dagger}$  & \textbf{34.03}$^{**\dagger}$  & \textbf{27.66}$^{**\dagger}$  & \textbf{34.55} \\
			\hline
		\end{tabular}
	}
	\scalebox{.95}{
		\begin{tabular}{l|c|c|c|c|c}
			\hline
			{\textbf{EN-DE}} & {\textbf{Dev (newstest2012)}} & {\textbf{newstest2013}} & {\textbf{newstest2014}} & {\textbf{newstest2015}} & {\textbf{AVG}}\\
			\hline
			PBSMT                    & 14.89     & 16.75    & 15.19    & 16.84    & 16.35  \\
			\hline
			GlobalAtt                & 17.09     & 20.24    & 18.67    & 19.78    & 19.56  \\
			Chen et al. (\citeyear{Chen-EtAl:2017:EMNLP2017})         & 17.48     & 21.03    & 19.43    & 20.56    & 20.31  \\
			LocalAtt                 & 17.19     & 20.74    & 19.00    & 20.15    & 19.96  \\
			FlexibleAtt              & 17.24     & 20.57    & 19.12    & 20.03    & 19.91  \\
			\hline
			\textbf{SDAtt}           & \textbf{17.86}   & \textbf{21.71}$^{**\dagger}$  & \textbf{20.36}$^{**\dagger}$   & \textbf{21.57}$^{**\dagger}$  & \textbf{21.21}  \\
			\hline
			\hline
		\end{tabular}
	}
	\caption{Results on ZH-EN and EN-DE translation tasks for the proposed SDAtt. ``*" indicates that the model significantly outperforms GlobalAtt at \emph{p}-value$<$0.05, ``**" indicates that the model significantly outperforms GlobalAtt at \emph{p}-value$<$0.01.
		``$\dagger$" indicates that the model significantly outperforms the best baseline Chen et al.\citeyear{Chen-EtAl:2017:EMNLP2017}'s Model at \emph{p}-value$<$0.05.
		\textbf{AVG} is the average BLEU score for all test sets.
		The bold indicates that the BLEU score of test set is better than the best baseline system.}
	\label{tab:ResultsSAttNMT}
\end{table*}

\begin{table*}[t!]
	\centering
	\scalebox{.95}{
		\begin{tabular}{l|c|c|c|c|c|c|c}
			\hline
			\hline
			{\textbf{ZH-EN}} & {\textbf{Dev (NIST02)}} & {\textbf{NIST03}} & {\textbf{NIST04}} & {\textbf{NIST05}} & {\textbf{NIST06}} & {\textbf{NIST08}} & {\textbf{AVG}}\\
			\hline
			PBSMT                      & 33.15   & 31.02   & 33.78  & 30.33   & 29.62   & 23.53   & 29.66 \\
			\hline
			GlobalAtt                  & 37.12   & 35.24   & 37.49  & 34.60   & 32.48   & 26.32   & 33.23  \\
			\quad +Chen et al. (\citeyear{Chen-EtAl:2017:EMNLP2017})   & 38.11   & 37.35   & 39.00   & 36.12    & 33.78   & 27.81   & 34.81  \\
			\quad +LocalAtt            & 37.89   & 37.06   & 38.73  & 36.10   & 33.62   & 27.43   & 34.59 \\
			\quad +FlexibleAtt         & 37.97   & 36.86   & 38.56  & 35.62   & 33.94   & 27.37   & 34.47 \\
			\hline
			\quad +\textbf{SDAtt}      & \textbf{38.61}   & \textbf{38.19}$^{**\dagger}$    & \textbf{39.81}$^{**\dagger}$   & \textbf{36.74}$**$   & \textbf{34.63}$^{**\dagger}$   & \textbf{28.61}$^{**\dagger}$     & \textbf{35.60} \\
			\hline
		\end{tabular}
	}
	\scalebox{.95}{
		\begin{tabular}{l|c|c|c|c|c}
			\hline
			{\textbf{EN-DE}} & {\textbf{Dev (newstest2012)}} & {\textbf{newstest2013}} & {\textbf{newstest2014}} & {\textbf{newstest2015}} & {\textbf{AVG}}\\
			\hline
			PBSMT                           & 14.89   & 16.75   & 15.19   & 16.84   & 16.35 \\
			\hline
			GlobalAtt                       & 17.09   & 20.24   & 18.67   & 19.78   & 19.56 \\
			\quad +Chen et al. (\citeyear{Chen-EtAl:2017:EMNLP2017})  & 18.03    & 21.44    & 19.96   & 21.07   & 20.82 \\
			\quad +LocalAtt                 & 17.78   & 21.26      & 19.87    & 20.67    &  20.6\\
			\quad +FlexibleAtt              & 17.56   & 21.10      & 19.76    & 20.74    & 20.53  \\
			\hline
			\quad +\textbf{SDAtt}           & \textbf{18.65}   & \textbf{22.11}$^{**\dagger}$   & \textbf{20.75}$^{**\dagger}$   & \textbf{22.05}$^{**\dagger}$   & \textbf{21.64} \\
			\hline
			\hline
		\end{tabular}
	}
	\caption{Results on ZH-EN and EN-DE translation tasks for the double context mechanism.}
	\label{tab:ResultsDCAttNMT}
\end{table*}
\subsection{Baseline Systems}
Along with the standard phrase-based SMT (\textbf{PBSMT}) implemented in Moses~\cite{koehn-EtAl:2007:PosterDemo} and standard NMT with global attention (\textbf{GlobalAtt})~\cite{Bahdanau-EtAl:2015:ICLR2015} baseline systems, we also compared the proposed methods to the recent related NMT methods:
\begin{itemize}
	\item \textbf{Chen et al.(\citeyear{Chen-EtAl:2017:EMNLP2017})}: extracted a local source dependency unit (including parent, siblings, and children of each source word) and learned its semantic representation.
	They introduced source dependency representation into the Encoder and Decoder by two kinds of NMT models, which  extended source word with dependency representation and enhanced the global attention with dependency representation, respectively. 
	Their methods are one of state-of-the-art syntax based NMT methods, which outperformed significantly the method of \cite{sennrich-haddow:2016:WMT}.
	\item \textbf{LocalAtt}: Luong et al.~\citeyear{luong-pham-manning:2015:EMNLP} selectively computed alignment probabilities for fixed-window source words centering around current aligned source position instead of all source words.
	\item \textbf{FlexibleAtt}: \citeauthor{shu-nakayama:2017:NMT}~\citeyear{shu-nakayama:2017:NMT} proposed a flexible attention NMT, which can dynamically create a window of the encoder states instead of fixed-window method of \cite{luong-pham-manning:2015:EMNLP}, and thus learned a flexible context to predict target word.
	\item \textbf{GlobalAtt+LocalAtt/FlexibleAtt}: we implemented the global attention with additional the local/flexible attention, to further evaluate our double context NMT.
\end{itemize}
\begin{figure*}[t]
	\centering
	\begin{minipage}[b]{0.48\linewidth}
		\centering
		\includegraphics[width=0.92\textwidth]{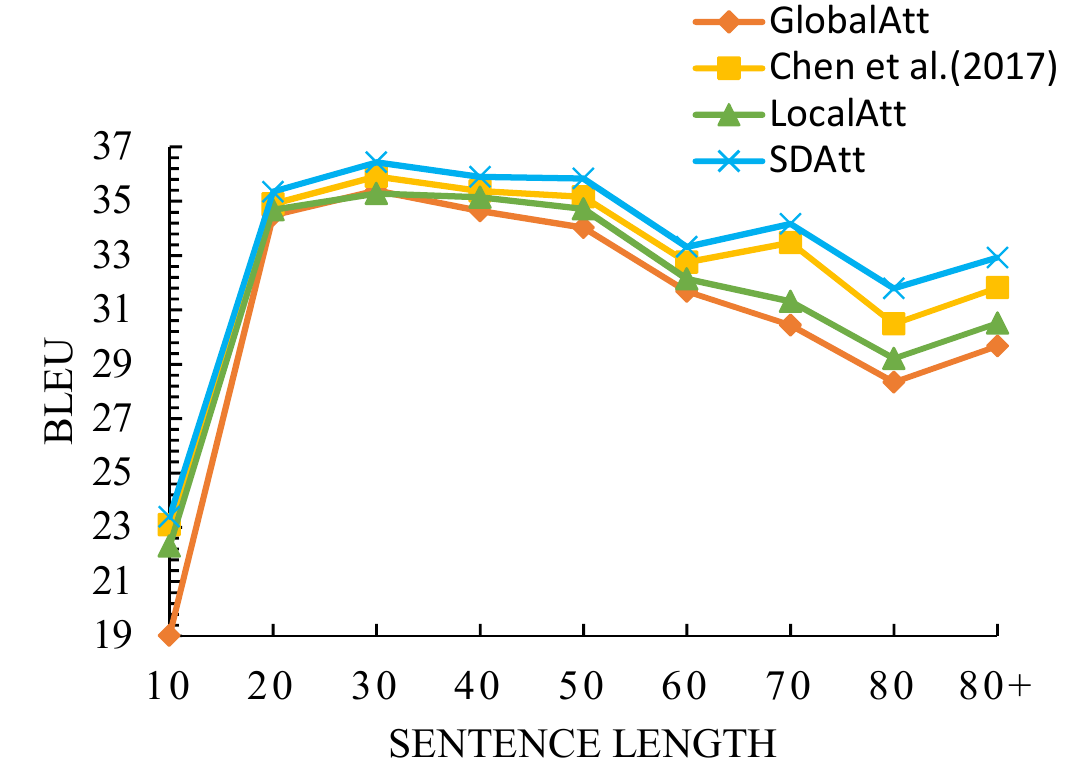}
		\caption{Translation qualities of different sentence lengths for SDAtt on the ZH-EN task.}
		\label{fig:ZHENSALength}
	\end{minipage}
	\centering
	\begin{minipage}[b]{0.48\linewidth}
		\centering
		\includegraphics[width=0.92\textwidth]{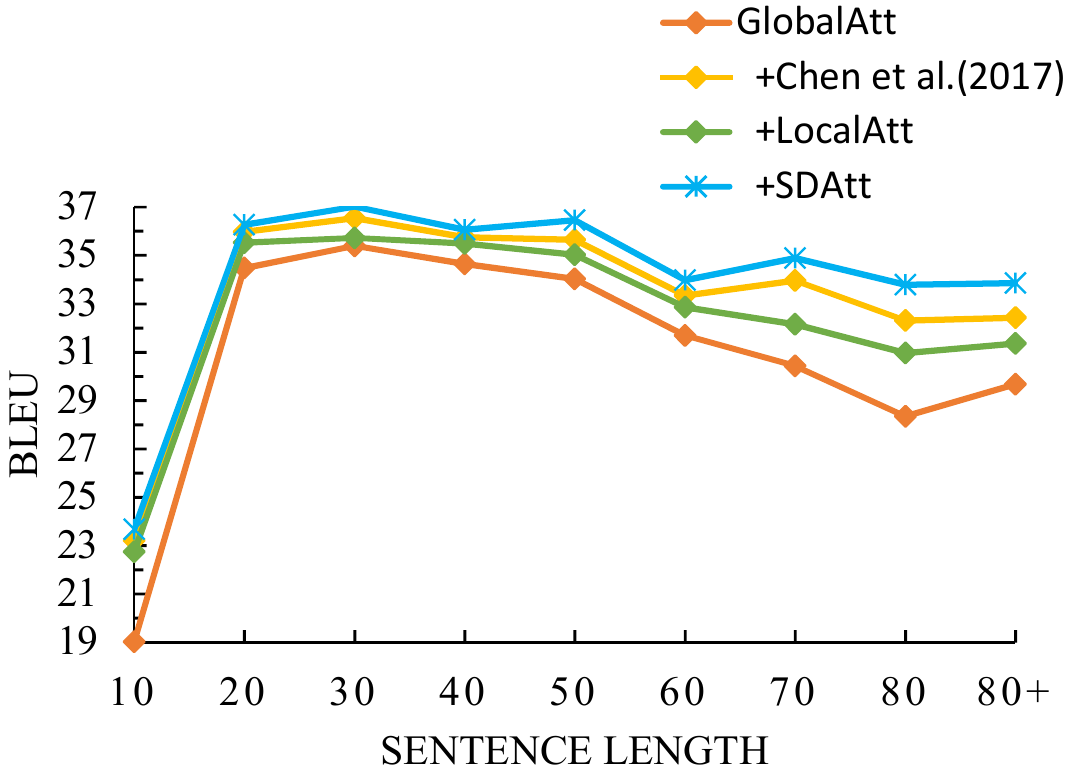}
		\caption{Translation qualities of different sentence lengths for GlobalAtt+SDAtt on the ZH-EN task.}
		\label{fig:ZHENDCLength}
	\end{minipage}
	\centering
	\begin{minipage}[b]{0.48\linewidth}
		\centering
		\includegraphics[width=0.92\textwidth]{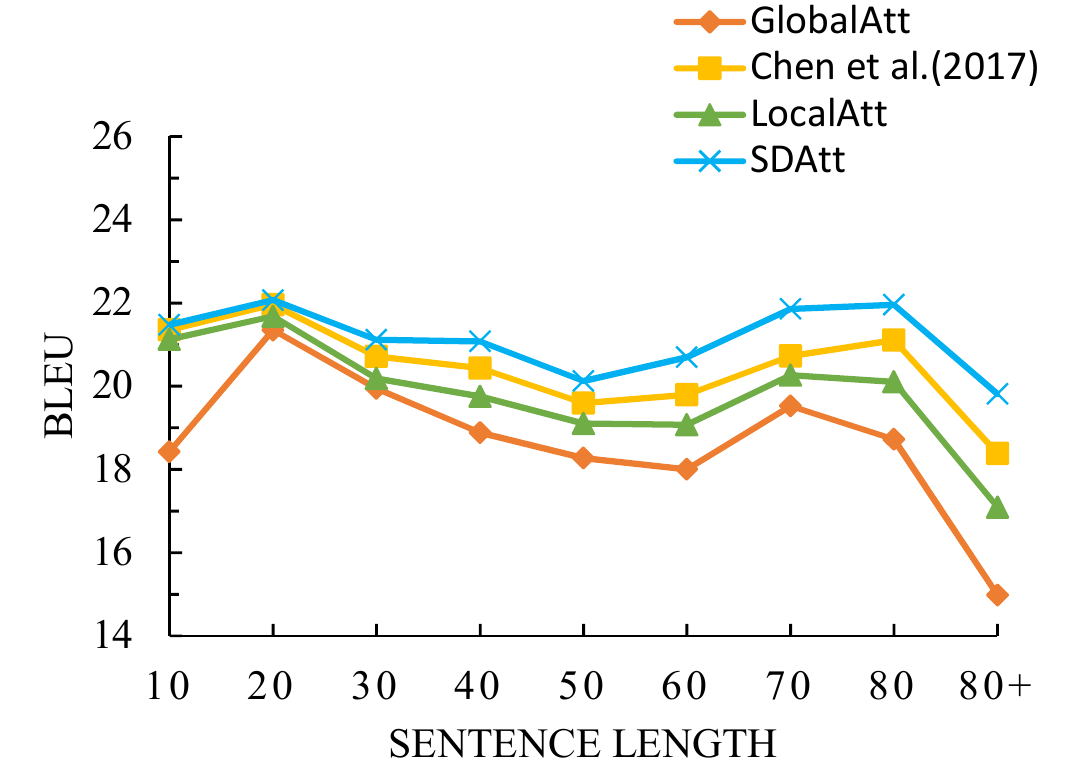}
		\caption{Translation qualities of different sentence lengths for SDAtt on the EN-DE task.}
		\label{fig:ENDESALength}
	\end{minipage}
	\centering
	\begin{minipage}[b]{0.48\linewidth}
		\centering
		\includegraphics[width=0.92\textwidth]{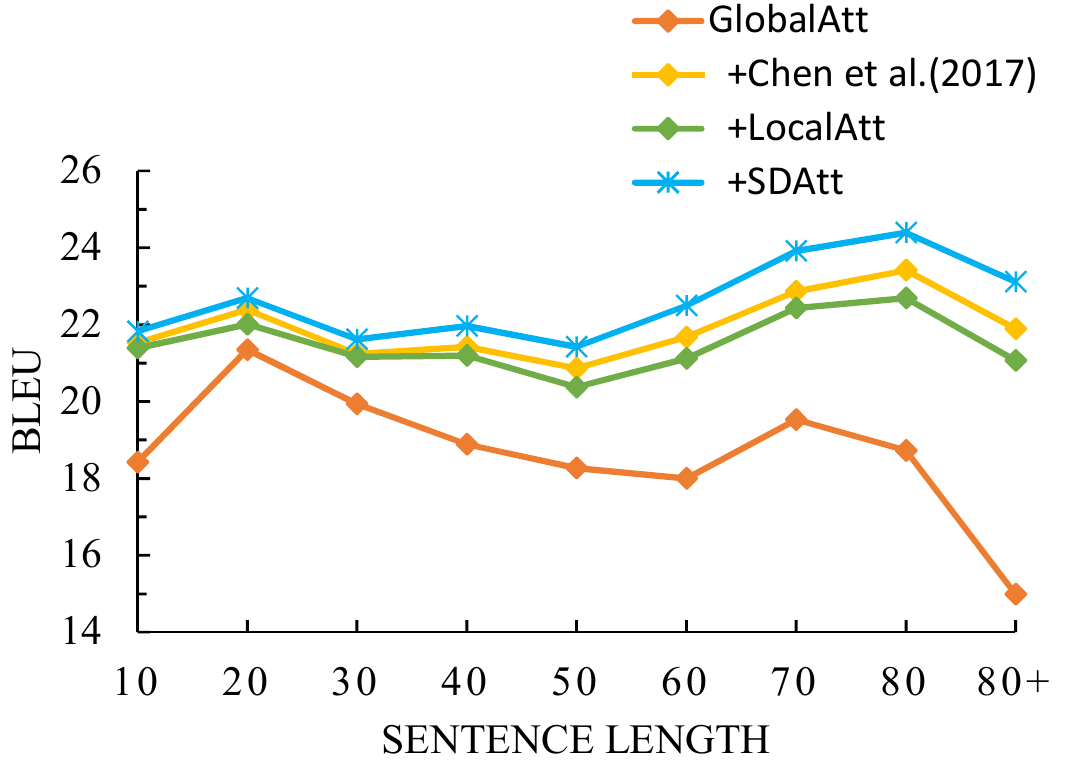}
		\caption{Translation qualities of different sentence lengths for GlobalAtt+SDAtt on the EN-DE task.}
		\label{fig:ENDEDCLength}
	\end{minipage}
\end{figure*}

All NMT models were implemented in the NMT toolkit Nematus~\cite{sennrich-EtAl:2017:EACLDemo}.\footnote{https://github.com/EdinburghNLP/nematus}
We used the Stanford parser~\cite{chang-EtAl:2009:SSST} to generate the dependency trees for source language sentences, such as Chinese sentences of ZH-EN and English sentences of EN-DE translation tasks.
We limited the source and target vocabularies to \emph{50}K, and the maximum sentence length was \emph{80}.
We shuffled training set before training and the mini-batch size is \emph{80}.
The word embedding dimension was \emph{620}-dimensions and the hidden layer dimension was \emph{1000}-dimensions, and the default dropout technique~\cite{DBLP:journals/corr/abs-1207-0580} in Nematus was used on the all layers.
Our NMT models were trained about \emph{400}k mini-batches using ADADELTA optimizer~\cite{DBLP:journals/corr/abs-1212-5701}, taking six days on a single Tesla P100 GPU, and the beam size for decoding was \emph{12}.
Case-insensitive \emph{4}-gram NIST BLEU score~\cite{papineni-EtAl:2002:ACL} was as the evaluation metric, and the signtest~\cite{collins-koehn-kucerova:2005:ACL} was as statistical significance test.

\subsection{Evaluating SDAtt NMT}
Table~\ref{tab:ResultsSAttNMT} shows translation results on ZH-EN and EN-DE translation tasks for syntax-directed attention NMT in Section~\ref{SAttNMT}.
The GlobalAtt significantly outperforms PBSMT by \emph{3.57} BLEU points on average, indicating that it is a strong baseline NMT system.
All the comparison methods, including Chen et al.(\citeyear{Chen-EtAl:2017:EMNLP2017})'s model, LocalAtt, and FlexibleAtt, outperform the baseline GlobalAtt.

(1) Over the GlobalAtt, the proposed SDAtt gains an improvement of 1.32 BLEU points on average on ZH-EN translation task, which indicates that our method can effective improve translation performance of NMT.

(2) The SDAtt surpasses LocalAtt and FlexibleAtt by 0.97/1.04 BLEU points on average on ZH-EN translation task.
This indicates that the proposed syntax distance constraint can capture more translation information to improve word prediction than linear distance constraint.

(3) The SDAtt also outperforms Chen et al.(\citeyear{Chen-EtAl:2017:EMNLP2017})'s model on ZH-EN translation task by 0.47 BLEU points on average.
This shows that our method can capture more source dependency information to improve word prediction.

(4) For EN-DE translation task, the proposed SDAtt gives similar improvements over the baseline system and comparison methods.
These results show that our method also can effectively improve the English-to-German translation task.
In other words, the proposed SDAtt is a robust method for improving the translation of other language pairs.

\subsection{Evaluating Double Context Mechanism}
To further verify the effectiveness of the proposed double context mechanism, we compared it with three similar models, including +Chen et al.~(\citeyear{Chen-EtAl:2017:EMNLP2017})'s Model, +LocalAtt, and +FlexibleAtt.
Table~\ref{tab:ResultsDCAttNMT} showed translation results of the proposed double context method on ZH-EN and EN-DE translation tasks.

(1) All the comparison methods and our +SDAtt outperform the baseline GlobalAtt.
In particularly, they gain further improvements by the corresponding single context NMT in Table~\ref{tab:ResultsSAttNMT}, for example, +FlexibleAtt (34.81) \emph{vs} LocalAtt (33.58).
This indicates that the proposed double-context mechanism for NMT is more effective than single context NMT.

(2) The +SDAtt outperforms GlobalAtt by 2.37 BLEU points on average on ZH-EN translation task.
Especially, the +SDAtt gains improvements of 1.01/1.13 BLEU points on average over the +LocalAtt/FlexibleAtt.
This shows that the proposed SDAtt give more translation information for NMT from source representation.

(3) The +SDAtt outperforms +Chen et al.(\citeyear{Chen-EtAl:2017:EMNLP2017})'s Model by 0.79 BLEU points on average on ZH-EN translation task.
This means that the SDAtt is more effective than enhancing global attention with source dependency representation of Chen et al. (\citeyear{Chen-EtAl:2017:EMNLP2017}).

(4) For the EN-DE translation task, the proposed +SDAtt shows similar improvements over the baseline system and comparison methods.
These results indicate that our double context architecture also can effectively improve the English-to-German translation task.

\subsection{Effect of Translating Long Sentences}
We grouped sentences of similar lengths on the test sets of the two tasks to evaluate the BLEU performance.
For example,  sentence length ``\emph{50}" indicates that the length of source sentences is between $40$ and $50$.
We then computed a BLEU score per group, as shown in Figures \ref{fig:ZHENSALength}-\ref{fig:ENDEDCLength}.

Take ZH-EN task as a example in Figure \ref{fig:ZHENSALength} and \ref{fig:ZHENDCLength}, our methods, including SDAtt and +SDAtt, always yielded consistently higher BLEU scores than the baseline GlobalAtt in terms of different lengths.
When the length came to ``30", they outperformed the best baseline Chen et al.~(\citeyear{Chen-EtAl:2017:EMNLP2017}).
This was because our methods can selectively focus on syntactic related source inputs with the current predicted target word and capture more source information to improve the performance of NMT.
Moreover, our models also showed similar improvements for EN-DE task in Figures \ref{fig:ENDESALength} and \ref{fig:ENDEDCLength}.
This again showed the effectiveness of our method on long sentence translation.

\section{Related Work}
\label{RWork}
Recently, many efforts have been initiated on exploiting source- or target-side syntax information to improve the performance of NMT.
\citeauthor{sennrich-haddow:2016:WMT} (\citeyear{sennrich-haddow:2016:WMT}) augmented each source word with its corresponding part-of-speech tag, lemmatized form and dependency label.
Li et al. (\citeyear{LiJunhui-EtAl:2017}) linearized parse trees of source sentences to obtain structural label sequences, thus capturing syntax label information and hierarchical structures.
To more closely combine the NMT with syntax tree, Eriguchi et al. (\citeyear{Eriguchi-EtAl:2017}) proposed a hybrid model that learns to parse and translate by combining the recurrent neural network grammar into the attention-based NMT, and thus encouraged the NMT model to incorporate linguistic prior during training, and lets it translate on its own afterward.
Wu et al. (\citeyear{wu-EtAl:2017:Long2}) then proposed a sequence-to-dependency NMT model, which used two RNNs to jointly generate target translations and construct their syntactic dependency trees, and then used them as context to improve word generation.
They extended source word with external syntax labels, thus providing richer context information for word prediction in NMT.

Eriguchi et al. (\citeyear{eriguchi-hashimoto-tsuruoka:2016:P16-1}) proposed a tree-to-sequence attentional NMT, which use a tree-based encoder to compute the representation of the source sentence following its parse tree instead of the sequential encoder. It further was extended by bidirectional tree encoder which learns both sequential and tree structured representations~\cite{Huadong-EtAl:2017}. Wu et al. (\citeyear{ijcai2017-584}) enriched each encoder state from both child-to-head and head-to-child with global knowledge from the source dependency tree.
Chen et al. (\citeyear{Chen-EtAl:2017:EMNLP2017}) extended each source word with local dependency unit to capture source long-distance dependency constraints, achieving an state-of-the-art performance in NMT, especially on long sentence translation.
These methods focused on enhancing source representation by capturing syntax structures in the source sentence or target sentence, such as phrase structures and dependency structures for improving translation.

In this paper, we extend the local attention with a novel syntax distance constraint to capture syntax related encoder states with the predicted target word.
Following the dependency tree of a source sentence, each source word has a distance mask, which denotes its syntax distances from the other source words. 
This mask is called as the syntax-distance constraint.
The decoder then focuses on the syntax-related source words within this syntax-distance constraint to compute a more effective context vector for predicting target word.
Moreover, we further propose a double context NMT architecture, which consists of a global context vector and a syntax-directed local context vector from the global attention, to provide more translation performance for NMT from source representation.

This work refines the local attention by syntax-distance constraint instead of traditional linear distance in the global or local attention, and thus selectively focuses on syntax-related source words to compute a more effective context vector for predicting target word.
\section{Conclusion}
\label{Conclusion}
In this paper, we explored the effect of syntactic distance on the attention mechanism.
We then proposed a syntax-directed attention for NMT method to selectively focus on syntax related source words for predicting target word.
Moreover, we further proposed a double context NMT architecture to provide more translation performance for NMT from source representation.
In the future, we will exploit richer syntax information to improve the performance of NMT.

\bibliography{emnlp-ijcnlp-2019}
\bibliographystyle{acl_natbib}

\end{document}